# Robust algorithm for calibration of robotic manipulator model

A. Klimchik*, Y. Wu*, G. Abba**, S. Garnier***, B. Furet*** and A. Pashkevich*

*Ecole des Mines de Nantes, 4 rue Alfred-Kastler, Nantes 44307, France; IRCCyN, Nantes 44000,France
(Tel: +33-251-85-83-00; e-mail: {alexandr.klimchik, yier.wu, anatol.pashkevich}@mines-nantes.fr).
**École Nationale d'Ingénieurs de Metz, Metz 57070, France; LCFC-ENSAM ParisTech Metz, (e-mail: abba@enim.fr).
***University of Nantes, France; IRCCyN, Nantes 44000,France, (e-mail:{sebastien.garnier, benoit.furet}@univ-nantes.fr)

**Abstract:** The paper focuses on the robust identification of geometrical and elastostatic parameters of robotic manipulator. The main attention is paid to the efficiency improvement of the identification algorithm. To increase the identification accuracy, it is proposed to apply the weighted least square technique that employs a new algorithm for assigning of the weighting coefficients. The latter allows taking into account variation of the measurement system precision in different directions and throughout the robot workspace. The advantages of the proposed approach are illustrated by an application example that deals with the elasto-static calibration of industrial robot.

## 1. INTRODUCTION

The problem of robot calibration is in the focus of the research community for many years (Stone 1987, Hollerbach 1989, Elatta 2004). However, there is a very limited number of works that directly address the issue of the identification accuracy and reduction of the calibration errors (Mooring 1991, Sun 2008, Hollerbach 2008). In general, there exist two main methods to improve the identification accuracy without increasing the number of experiments. The first of them consists in optimization of the manipulator measurement configuration used for the calibration experiments. The second method deals with proper tuning of the identification algorithms used for estimation of the manipulator parameters (fine choosing of the weighting coefficients, for instant). As follows from the literature analysis, the first method has been studied in a number of papers (Khalil 1991, Borm 1991, Daney 2002), while the second one received less attention in robotic research. For this reason, taking into account particularities of the measurement system used in our experiments, this paper focuses on the enhancement of the second method, which looks rather promising here.

To identify the desired parameters, most of the robot calibration procedure employ the ordinary least-square technique, where all identification equations are treated similarly, with the same weights. This approach perfectly suits to the measurement systems that provide roughly the same precision in all directions and in all workspace points. Mathematically, it corresponds to the i.i.d.-hypothesis concerning the measurement noise (i.e. to the assumption that all measurement errors are unbiased, independent and identically distributed). However, in this study, at least one of these assumptions is violated because the precision of the laser-tracker used in the calibration experiments essentially depends on the direction and the target marker location in the manipulator workspace.

To overcome this difficulty, the weighted least-square technique can be applied. As known from literature, for the linear regression it gives rather good improvement and allows essentially reducing the measurement errors impact. It is evident that for the robotic calibration problem, similar benefits can be gained, but the weighting coefficient selection is non-trivial here because of the high non-linearity of the equations describing the robotic manipulator.

To address this problem, the remainder of this paper is organized as follows. Section 2 defines the research problem and also presents some experimental data concerning statistical properties of the measurement errors. In Sections 3, the identification algorithm is presented that is based on the weighted least-square. Section 4 deals with selection of the weighting coefficients that ensure robustness of the identification algorithm with respect to the measurement noise. Section 5 presents an application example, which illustrates benefits of the proposed approach. And finally, Section 6 summarizes the main contributions of the paper.

## 2. PROBLEM STATEMENT

In robotics, the calibration procedure can be treated as the best fitting of the experimental data using corresponding manipulator model

$$[\Delta\mathbf{\Pi}, \mathbf{k}] = \arg\min\left(\mathbf{p}_i - f(\mathbf{L}, \mathbf{q}_i, \Delta\mathbf{\Pi}, \mathbf{k}, \mathbf{p}_i^0, \mathbf{F}_i)\right) \quad (1)$$

which describes its geometrical and elastostatic behavior defined by the known function $f(.)$ whose parameters should be tuned. Here $\mathbf{p}_i$ is the vector of measurements (the Cartesian coordinates of the end-effector target points), the vector **L** collects all geometrical parameters, $\mathbf{q}_i$ is the vector of actuated coordinates, the vector $\Delta\mathbf{\Pi}$ collects errors in geometrical parameters, the vector **k** collects all compliances of the manipulator, the vectors $\mathbf{p}_i^0$ and $\mathbf{F}_i$ are used for elastostatic calibration only and correspond to the measurement position without loading and applied external loading respectively.

In usual engineering practice it is assumed that all measurements ($\mathbf{p}_i$ and $\mathbf{p}_i^0$) are corrupted by the same measurement noise $N(0,\sigma^2)$, which induces errors $\boldsymbol{\varepsilon}_i$ with zero expectation $E(\boldsymbol{\varepsilon}_i) = \mathbf{0}$ and diagonal covariance matrix $E(\boldsymbol{\varepsilon}_i\boldsymbol{\varepsilon}_i^T) = \sigma^2 \cdot \mathbf{I}$. However, for many measurement devices such as laser-trackers, the precision highly depends on the measurement direction and vary throughout the robot workspace. In this case, the covariance matrix can be still assumed to be diagonal, but with non-equal principle elements, i.e. $E(\boldsymbol{\varepsilon}_i\boldsymbol{\varepsilon}_i^T) = diag(\sigma_{xi}^2, \sigma_{yi}^2, \sigma_{zi}^2)$, where $\sigma_{xi}$, $\sigma_{yi}$, $\sigma_{zi}$ are different and vary from measurement to measurement. This phenomena is illustrated by experimental data presented in Table 1, which includes dispersions of the measurement errors for the Cartesian coordinates x, y, z for several measurement configurations used in conventional calibration experiments. These results has been obtained by processing the measurement data for 15 configurations and 18 independent experiments for each of them (i.e. by using data set which consists of 810 values). It should be noted that the manipulator repeatability, which is about 60μm, does not have influence on the presented results because of specificity of the measurement experiments, where only difference in the Cartesian coordinate variations were evaluated (before and after applying external loading). As follows from the presented results, the measurement error dispersion vary from 17μm to 153μm and highly depends both on the direction (x, y or z) and the end-effector location in the manipulator workspace (corresponding to the configurations #1-#15). In particular, for the same configuration $\sigma_{xi}$, $\sigma_{yi}$, $\sigma_{zi}$ can differ by the factor 5. Besides, from one configuration to another corresponding dispersions may differ by 7 times.

**Table 1. Dispersions of measurement errors in deflections for different test configurations**

| Configuration | $\sigma_x$, [μm] | | $\sigma_y$, [μm] | | $\sigma_z$, [μm] | |
|---|---|---|---|---|---|---|
| | mean | std | mean | std | mean | std |
| Conf. #1 | 150 | ±1 | 64 | ±1 | 33 | ±1 |
| Conf. #2 | 57 | ±4 | 86 | ±8 | 118 | ±15 |
| Conf. #3 | 97 | ±9 | 70 | ±5 | 44 | ±8 |
| Conf. #4 | 28 | ±1 | 19 | ±1 | 35 | ±1 |
| Conf. #5 | 72 | ±3 | 48 | ±4 | 17 | ±1 |
| Conf. #6 | 153 | ±8 | 46 | ±3 | 22 | ±1 |
| Conf. #7 | 112 | ±6 | 66 | ±3 | 53 | ±4 |
| Conf. #8 | 74 | ±5 | 55 | ±3 | 59 | ±1 |
| Conf. #9 | 80 | ±9 | 63 | ±7 | 102 | ±15 |
| Conf. #10 | 69 | ±2 | 73 | ±1 | 79 | ±1 |
| Conf. #11 | 80 | ±3 | 36 | ±1 | 26 | ±3 |
| Conf. #12 | 53 | ±4 | 39 | ±1 | 29 | ±1 |
| Conf. #13 | 26 | ±1 | 29 | ±1 | 29 | ±1 |
| Conf. #14 | 88 | ±4 | 121 | ±1 | 42 | ±1 |
| Conf. #15 | 90 | ±6 | 52 | ±3 | 50 | ±1 |

Hence, usual assumptions incorporating in robot calibration techniques concerning measurement noise properties should be revised. This motivates the principle goal of the paper that is aimed at developing of the robust identification algorithm that takes into account variations of $\sigma_{xi}$, $\sigma_{yi}$, $\sigma_{zi}$.

## 3. IDENTIFICATION ALGORITHM

In manipulator geometric calibration, the basic expression i usually written as follows

$$\Delta\mathbf{p}_i^g = \mathbf{J}_i^{(p)}(\mathbf{L}, \mathbf{q}_i) \cdot \Delta\mathbf{\Pi} \qquad (2)$$

where $\Delta\mathbf{p}_i^g$ is the difference between the computed via direct geometrical model and measured end-effector position, the matrix $\mathbf{J}_i^{(p)}$ is the geometrical Jacobian and the superscript '(p)' specifies only components that are related to the robot position. It is clearly the linearized model, but it is valid here since in practice the geometrical errors are low enough compared to the nominal values of the manipulator parameters.

Similarly, for the elastostatic calibration, the basic expression can be presented in the form

$$\Delta\mathbf{p}_i^e = \mathbf{A}_i^{(p)}(\mathbf{L}, \mathbf{q}_i, \mathbf{F}_i) \cdot \mathbf{k} \qquad (3)$$

where $\Delta\mathbf{p}_i^e$ is the vector of the end-effector displacements under the loading $\mathbf{F}_i$, the vector **k** collects all compliances of the manipulator, the matrix $\mathbf{A}_i^{(p)}$ defines the mapping between the joint compliances **k** and the and-effector displacements. For the further convenience, both expression (2), (3) can be integrated in a single one

$$\Delta\mathbf{p}_i = \mathbf{B}_i^{(p)}(\mathbf{L}, \mathbf{q}_i, \mathbf{F}_i) \cdot \mathbf{X} \qquad (4)$$

where $\Delta\mathbf{p}_i$ is the vector of measurements and $\mathbf{X} = [\Delta\mathbf{\Pi}, \mathbf{k}]$ the vector of the unknown parameters that should be identified, $\mathbf{B}_i^{(p)}$ is corresponding matrix function.

Using above defined notations, the calibration can be reduced to the following optimization problem

$$F = \sum_{i=1}^{m}(\mathbf{B}_i^{(p)}\mathbf{X} - \Delta\mathbf{p}_i)^T(\mathbf{B}_i^{(p)}\mathbf{X} - \Delta\mathbf{p}_i) \to \min_{\mathbf{X}} \qquad (5)$$

that can be solved using the least square approach. It leads to the following solution

$$\hat{\mathbf{X}} = \left(\sum_{i=1}^{m}\mathbf{B}_i^{(p)T}\mathbf{B}_i^{(p)}\right)^{-1} \cdot \left(\sum_{i=1}^{m}\mathbf{B}_i^{(p)T}\Delta\mathbf{p}_i\right) \qquad (6)$$

If the measurement noise is Gaussian (as it is assumed in conventional calibration techniques) the covariance matrix for the parameter estimates $\hat{\mathbf{X}}$ can be computed as follows

$$cov(\hat{\mathbf{X}}) = \sigma^2\left(\sum_{i=1}^{m}\mathbf{B}_i^{(p)T}\mathbf{B}_i^{(p)}\right)^{-1} \qquad (7)$$

where $\sum_{i=1}^{m}\mathbf{B}_i^{(p)T}\mathbf{B}_i^{(p)}$ is the so-called information matrix.

However, if the measurement noise varies from configuration to configuration the previous expression should be revised. It

can be proved that in general case the covariance matrix is expressed as

$$\text{cov}(\hat{\mathbf{X}}) = \left(\mathbf{B}_a^{(p)T} \cdot \mathbf{B}_a^{(p)}\right)^{-1} \left(\mathbf{B}_a^{(p)T} \cdot \mathbf{\Sigma}^2 \cdot \mathbf{B}_a^{(p)}\right) \left(\mathbf{B}_a^{(p)T} \cdot \mathbf{B}_a^{(p)}\right)^{-1} \quad (8)$$

where the matrix $\mathbf{\Sigma} = diag(\sigma_1, \sigma_2, ..., \sigma_{3m})$ describes the statistical properties of the measurement errors and the matrix $\mathbf{B}_a^{(p)}$ aggregates all matrices $\mathbf{B}_i^{(p)}$ from expression (4).

In order to improve the identification accuracy, it is reasonable to modify the objective function (5) and rewrite it in the form

$$F = \sum_{i=1}^m (\mathbf{B}_i^{(p)}\mathbf{X} - \Delta\mathbf{p}_i)^T \mathbf{W}^2 (\mathbf{B}_i^{(p)}\mathbf{X} - \Delta\mathbf{p}_i) \to \min_{\mathbf{X}} \quad (9)$$

where $\mathbf{W} = diag(w_1, w_2, ..., w_{3m})$ is the matrix of weighting coefficients. This leads to slightly different expression for the parameter estimates

$$\hat{\mathbf{X}} = \left(\mathbf{B}_a^{(p)T} \cdot \mathbf{W}^2 \cdot \mathbf{B}_a^{(p)}\right)^{-1} \cdot \left(\mathbf{B}_a^{(p)T} \cdot \mathbf{W}^2 \cdot \Delta\mathbf{p}_a\right) \quad (10)$$

where $\Delta\mathbf{p}_a$ aggregates $\Delta\mathbf{p}_i$. In this case, the covariance matrix can be computed as

$$\text{cov}(\hat{\mathbf{X}}) = \left(\mathbf{B}_a^{(p)T} \mathbf{W}^2 \mathbf{B}_a^{(p)}\right)^{-1} \times \\ \times \mathbf{B}_a^{(p)T} \mathbf{W}^2 \mathbf{\Sigma}^2 \mathbf{W}^2 \mathbf{B}_a^{(p)} \left(\mathbf{B}_a^{(p)T} \mathbf{W}^2 \mathbf{B}_a^{(p)}\right)^{-1} \quad (11)$$

and, as follows from detailed analysis, the best selection of the weighting matrix corresponds to the equation $\mathbf{W} \cdot \mathbf{\Sigma} = \mathbf{I}$.

Hence, to assign the weighting coefficients, the measurement noise variance should be known. However, in practice, exact values of $\sigma_{xi}, \sigma_{yi}, \sigma_{zi}$ are unknown and the estimates can be used only. On the other hand, as follows from practical experience, small variations in the weighting coefficients are not critical and they do not affect significantly the identification accuracy. Nevertheless, if the weights are assigned using small number of experiments, the identification results can be unpredictably affected. Therefore, the problem of computing of the weighting coefficients that are able to ensure robustness of the identification algorithm is important here and it will be in the focus of next Section.

## 4. ASSIGNING WEIGHTING COEFFICIENTS

It can be proved that for the linear model, the weighing coefficients ensuring the lowest variance of the unknown parameters can be computed as

$$w_i = a / \sigma_i \quad (12)$$

where $\sigma_i$ is the standard deviation of the measurement noise for the $i$-th identification expression and the constant $a$ is the scalier factor introduced to avoid the problem of the units. For example, for the linear regression with a single scalar parameter, this approach allows to reduce the variance from $(\sum x_i^2 \sigma_i^2) / (\sum x_i^2)^2$ to $1 / \sum x_i^2 / \sigma_i^2$

Applying this idea to the problem of robot calibration, it is possible to transform the covariance matrix expression (11) to the form

$$\text{cov}(\hat{\mathbf{X}}) = \left(\mathbf{B}_a^{(p)T} \cdot \mathbf{\Sigma}^2 \cdot \mathbf{B}_a^{(p)}\right)^{-1} \quad (13)$$

where all elements are essentially lower compared to (8). Hence the problem of interests is to obtain $\sigma_i$ for each experiment (and for each coordinate), which will be used for computing weighting coefficients $w_i$.

As it was mentioned before, computing $\sigma_i$ is based on a few measurement may have the opposite effect - decrease identification accuracy. But this remark does not refer to our case since we have a group of points (18) in the vicinity of single robot configuration which allows us to estimate s.t.d. of measurement noise for x-, y- and z-directions with quite good precision (see Table 1). It should be noted that obtained values of $\sigma_i$ are considerably higher than the claimed accuracy of the measuring system (10 µm). Nevertheless, this value can be used as a normalized coefficient $a$. Besides, since in real experiments it is not possible to be insured against poor distribution of the measurement errors, in order to increase robustness of the identification algorithm, it is proposed to introduce steady component in the $\sigma_i$ that in practice can be assigned by the claimed precision of measurement system $\sigma_0 = 10\mu m$. Finally. expression for the weights takes a form

$$w_i = \sigma_0 / (\sigma_0 + \lambda \cdot \sigma_i) \quad (14)$$

where $\lambda$ is a scalar factor that allows to tune the impact of $\sigma_i$, $\sigma_0$ is normalization factor that also allows to avoid division by zero.

## 5. APPLICATION EXAMPLE

The developed calibration algorithm has been applied to the industrial robot KR-270. To take into account the compensator influence while remaining approach developed for serial robots without compensators (Pashkevich 2011), an equivalent virtual spring with non-linear stiffness depending on the joint variable $q_2$ is used. Using this idea, it is convenient to consider several independent parameters $k_{2i}$ corresponding to each value of $q_2$. So, the set of desired parameters $(k_{21}, k_{22}...), k_3, ..., k_6$ can be denoted as the vector $\mathbf{k}$. This allows us to obtain linear form of the identification equations.

To find optimal measurement configurations, the design of experiments has been carried out using industry-oriented performance measure proposed in (Klimchik 2012) for five different angles $q_2$ that are distributed between the joint limits. For each $q_2$ three optimal measurement configurations have been found taking into account physical constraints that are related to the joint limits and the possibility to apply the gravity force (work-cell obstacles and safety reasons). The results of the calibration experiment design are presented in Table 2.

**Table 2. Measurement configurations**

| Joint angles, [deg] | | | | | |
|---|---|---|---|---|---|
| $q_1$ | $q_2$ | $q_3$ | $q_4$ | $q_5$ | $q_6$ |
| 79.20 | -0.01 | -5.57 | 51.00 | -97.52 | -91.67 |
| 63.00 | | -12.22 | -56.49 | 41.42 | 150.55 |
| 63.00 | | -47.98 | -70.04 | -61.55 | 177.16 |
| 95.00 | -25.24 | 33.00 | 129.69 | -98.10 | 90.57 |
| 95.00 | | -107.01 | 109.95 | -61.19 | 174.21 |
| 105.00 | | 14.30 | 55.21 | 41.26 | -152.97 |
| 56.60 | -56.9 | 44.54 | -55.11 | 41.90 | 152.06 |
| 56.60 | | 64.73 | -129.65 | -98.260 | -90.55 |
| 144.80 | | 104.49 | -69.41 | 61.67 | -6.33 |
| -41.00 | -99.85 | -91.68 | 55.12 | 41.53 | -152.48 |
| -143.00 | | -32.64 | 110.31 | -61.47 | -6.29 |
| -143.00 | | -72.01 | 129.65 | -98.09 | 90.82 |
| 133.00 | -140 | 147.68 | 129.64 | -97.90 | 90.99 |
| -60.00 | | 7.59 | -110.09 | -61.36 | -174.09 |
| -60.00 | | -52.00 | -124.89 | -41.62 | 27.78 |

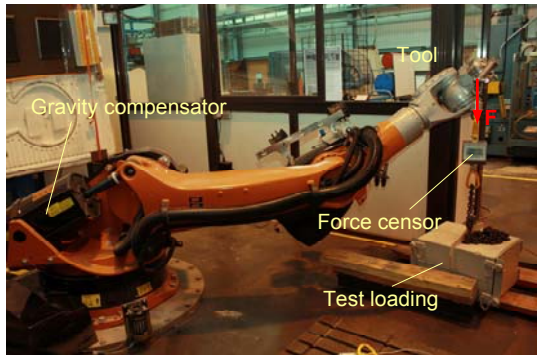

Fig. 1. Experimental setup for the identification of the elastostatic parameters

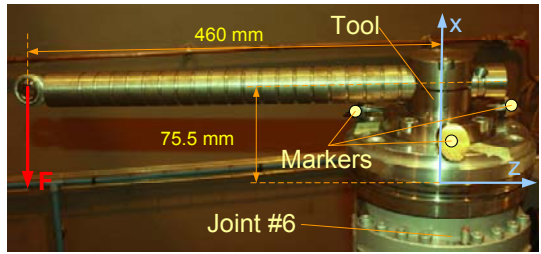

Fig. 2. End-effector used for elastostatic calibration experiments

To generate elastostatic deflections, the gravity forces have been applied to the robot end-effector (see Fig 1) using specific calibration tool (see Fig. 2). To obtain the desired set of initial data, the manipulator sequentially passed through several measurement configurations. Using the laser tracker with the claimed precession $10\mu m$, the Cartesian coordinates of the reference points have been measured twice, before and after loading. To increase identification accuracy, three reference points (markers) have been used and the loading of the maximum allowed magnitude 250-280 kg have been applied. In addition, to ensure high identification accuracy for each configuration, the experiments were repeated six times. In total the experimental data include 270 measurements which give 810 equations for identification of 9 desired parameters.

The identification has been performed using Ordinary Least-Square (OLS) and Weighted Least Square (WLS) techniques. For the second approaches the weights have been obtained using non-compensated deflections after identification joint compliances, normalization factor $\sigma_0$ has been assigned to the claimed precession of the laser tracker ($10\mu m$). Corresponding values of the elastostatic parameters are presented in Table 3, where for WLS the results for $\lambda = 0.5, 1, 2$ are proposed. It also includes the confidence intervals computed as $\pm 3\sigma$, where the standard deviation $\sigma$ has been evaluated based on the experimental data using expressions (8) and (13). The results show that the confidence intervals for OLS and WLS have intersections for all parameters of interest, moreover, the confidence intervals for WLS are always inside confidence intervals for OLS and considerably lower (see Table 4). Another important conclusion is that the choice of the coefficient $\lambda$ does not influence on the final results and for simplicity can be assigned to one.

**Table 3. Identified values of manipulator elasto-static parameters using different approaches, [rad/N×µm]**

| $k_i$ | OLS | WLS | | |
|---|---|---|---|---|
| | | $\lambda=0.5$ | $\lambda=1.0$ | $\lambda=2.0$ |
| $k_{21}$ | 0.297 ±0.010 | 0.287 ±0.0003 | 0.287 ±0.0003 | 0.287 ±0.0003 |
| $k_{22}$ | 0.287 ±0.012 | 0.277 ±0.0004 | 0.277 ±0.0004 | 0.277 ±0.0004 |
| $k_{23}$ | 0.315 ±0.018 | 0.302 ±0.0005 | 0.302 ±0.0005 | 0.302 ±0.0005 |
| $k_{24}$ | 0.302 ±0.032 | 0.293 ±0.0010 | 0.293 ±0.0010 | 0.293 ±0.0010 |
| $k_{25}$ | 0.251 ±0.020 | 0.246 ±0.0007 | 0.246 ±0.0007 | 0.246 ±0.0007 |
| $k_3$ | 0.396 ±0.031 | 0.416 ±0.0011 | 0.416 ±0.0011 | 0.416 ±0.0011 |
| $k_4$ | 3.017 ±0.248 | 2.786 ±0.0071 | 2.786 ±0.0071 | 2.786 ±0.0071 |
| $k_5$ | 3.294 ±0.506 | 3.483 ±0.0120 | 3.483 ±0.0120 | 3.483 ±0.0120 |
| $k_6$ | 2.248 ±0.725 | 2.074 ±0.0267 | 2.074 ±0.0267 | 2.074 ±0.0267 |

It should be noted that the best weighing coefficients can be computed iteratively, where starting from the second iterations the residuals have been computed for the elastostatic parameters identified using WLS with the weights obtained on the previous iteration. This allows us to increase additionally identification accuracy by the factor 3-20 comparing with single iteration WLS. Finally, the identification accuracy have been increased by the factors 27-42 and the errors do not overcome 1.23% for the $k_6$ and is 0.09-0.34%. for the remainder parameters of interest (such non-equivalent distribution of the identification errors is caused by the specific of design of calibration experiment, where the position accuracy after compensation of the elastic deflections has been chosen as a performance measure (Klimchik 2012)). The benefits of WLS are summarized in Table 4. It includes mutual locations of the results for OLS and WLS, identification accuracy and the benefits of new identification algorithm comparing with old one. The convergence of the iterative procedure presented in Figure 3. Figure 3a shows variation in $k_{21}$ from iteration to iteration and Figure 3b provides corresponding Confidence intervals. This results shows that parameter value does not change significantly (3.3% from the original one), however its precision has been increased by the factor 40. After 10 iterations the values of $k_{21}$ and CI are stabilized

**Table 4. Benefits of WLS**

| $k_i$ | Mutual location of CI | $CI_{OLS}$ | $CI_{WLS}$ | $\sigma_{OLS}/\sigma_{WLS}$ |
|---|---|---|---|---|
| $k_{21}$ | | 3.4% | 0.09% | 40.5 |
| $k_{22}$ | | 4.2% | 0.13% | 33.2 |
| $k_{23}$ | | 5.9% | 0.18% | 33.9 |
| $k_{24}$ | | 10.7% | 0.33% | 33.1 |
| $k_{25}$ | | 7.8% | 0.27% | 30.1 |
| $k_3$ | | 7.8% | 0.26% | 28.8 |
| $k_4$ | | 8.2% | 0.25% | 35.1 |
| $k_5$ | | 15.3% | 0.34% | 42.0 |
| $k_6$ | | 32.2% | 1.28% | 27.2 |

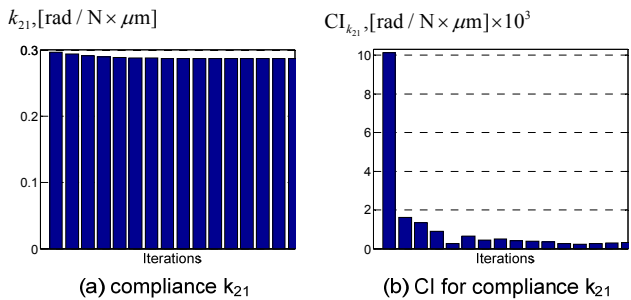

(a) compliance $k_{21}$     (b) CI for compliance $k_{21}$

Fig. 3. Convergences of identified value for $k_{21}$ and its CI

Hence using new identification algorithm the elasto-static parameters of the industrial robot Kuka KR-270 have been identified with a higher accuracy. This allows us to increase the efficiency of the error compensation technique proposed in (Klimchik 2013).

## 6. CONCLUSIONS

The paper presents a robust identification algorithm for geometrical and elastostatic calibration of robotic manipulator. In contrast to previous works, the proposed algorithm is based on the weighted least-square technique that employs a new strategy for assigning of the weighting coefficients. Such approach allows us to take into account variation of the measurement system precision in different directions and throughout the robot workspace. Application of this technique leads to the essential improvement of the identification accuracy. The advantages of the proposed approach are illustrated by an application example that deals with the elasto-static calibration of an industrial robot which is used for milling of large-dimensional composite parts. It shows that the proposed algorithm allows to reduce most of the identification errors down to 0.3%, which is 30-40 times better compared to the conventional technique.

## ACKNOWLEDGMENT

The work presented in this paper was partially funded by the ANR, France (Project ANR-2010-SEGI-003-02-COROUSSO). The authors also thank Fabien Truchet, Guillaume Gallot and Joachim Marais for their great help with the experiments.

## REFERENCES

Borm J., Menq C. (1991). Determination of optimal measurement configurations for robot calibration based on observability measure. *International Journal of Robotics Research*, vol. 10, no. 1, pp. 51–63.

Daney D. (2002). Optimal measurement configurations for Gough platform calibration. *IEEE International Conference on Robotics and Automation (ICRA 2002)*, pp. 147-152.

Elatta A.Y., Gen L.P., Zhi F.L., Daoyuan Y., Fei L. (2004). An Overview of Robot Calibration. *Information Technology Journal*, vol. 3, no. 1, pp. 74-78.

Hollerbach J. M. (1989). A Survey of Kinematic Calibration. *The Robotics Review 1,* Cambridge, MIT Press, pp. 207-242.

Hollerbach J.M., Khalil W., Gautier M. (2008). Model Identification, *Springer handbook of robotics, Part B*, pp 321-344

Khalil W., Gautier M., Enguehard Ch. (1991). Identifiable parameters and optimum configurations for robots calibration. *Robotica*, vol. 9, pp. 63-70.

Klimchik A., Wu Y., Pashkevich A., Caro S., Furet B. (2012). Optimal Selection of Measurement Configurations for Stiffness Model Calibration of Anthropomorphic Manipulators. *Applied Mechanics and Materials*, vol. 162, pp. 161-170

Klimchik A., Pashkevich A., Chablat D., Hovland G. (2013). Compliance error compensation technique for parallel robots composed of non-perfect serial chains. *Robotics and Computer-Integrated Manufacturing Journal*, vol. 29, issue 2, pp. 385–393.

Mooring B., Driels W.M., Roth Z. (1991). *Fundamentals of Manipulator Calibration*. John Wiley & Sons, NY, 1991.

Pashkevich A., Klimchik A., Chablat D. (2011). Enhanced stiffness modeling of manipulators with passive joints. *Mechanism and Machine Theory*, vol. 46, issue 5, pp. 662-679

Stone H. W. (1987). *Kinematic Modeling, Identification, and Central of Robot Forward Manipulators*, Boston, Kluwer.

Sun Y., Hollerbach J.M. (2008). Active robot calibration algorithm, *IEEE International Conference on Robotics and Automation (ICRA 2008)*, pp. 1276-1281.